%% file: main.tex
\documentclass[10pt,twocolumn,letterpaper]{article}

\usepackage[pagenumbers]{iccv} 

\usepackage{tikz}

\usepackage{times}
\input{setup/package}
\input{setup/macros}

\definecolor{bittersweet}{rgb}{1.0, 0.44, 0.37}

\definecolor{baselinecolor}{gray}{.9}

\definecolor{cadmiumgreen}{rgb}{0.0, 0.42, 0.24}

\newcommand{\ourmodel}{\textsc{AvED} }
\newcommand{\ourmodela}{\textsc{AvED}}

\newcommand{\ourdata}{\textsc{AvED}-Bench }
\newcommand{\ourdataa}{\textsc{AvED}-Bench}

\title{Zero-Shot Audio-Visual Editing via Cross-Modal Delta Denoising}

\author{
Yan-Bo Lin$^{1}$\thanks{Work done during an internship at Microsoft.}\quad\quad
Kevin Lin$^{2}$\quad\quad
Zhengyuan Yang$^{2}$\quad\quad 
Linjie Li$^{2}$\quad\quad 
Jianfeng Wang$^{2}$\quad\quad \\
Chung-Ching Lin$^{2}$\quad\quad 
Xiaofei Wang$^{2}$\quad\quad 
Gedas Bertasius$^{1}$\quad\quad
Lijuan Wang$^{2}$
\\
$^{1}$UNC Chapel Hill\quad 
$^{2}$Microsoft 
}

\begin{document}
\maketitle

\input{0_Abstract}
\input{1_intro}

\input{2_related_work}
\input{3_Method}
\input{4_result}
\input{5_Conclusion}

\newcount\cvprrulercount
\appendix
\input{6_appendix}
\clearpage
{\small
\bibliographystyle{ieee_fullname} 
\bibliography{egbib}
}

\end{document}

%% file: setup/package.tex
\usepackage{comment}

\usepackage{graphicx}
\usepackage{subcaption}
\usepackage{float}

\usepackage{booktabs}                   
\usepackage{multirow}
\usepackage{array}
\usepackage{enumitem}
\usepackage{paralist}

\usepackage{mathtools}
\usepackage{amssymb,amsmath}   

\usepackage{pifont}
\usepackage{xspace}

\usepackage{diagbox}
\usepackage{xcolor}
\usepackage[export]{adjustbox}
\usepackage{sidecap}

\usepackage{multirow, makecell, booktabs}
\definecolor{citecolor}{HTML}{0071bc}
\definecolor{linkcolor}{HTML}{ED1C24}

\usepackage[breaklinks=true,bookmarks=false,colorlinks,bookmarks=false, citecolor=citecolor,linkcolor=linkcolor, pagebackref]{hyperref}

%% file: setup/macros.tex
\def\eg{e.g.,~}               
\def\ie{i.e.,~}               

\newcommand{\figref}[1]{Figure~\ref{fig:#1}} 
\newcommand{\tabref}[1]{Table~\ref{tab:#1}}
\newcommand{\eqnref}[1]{Eq.~\ref{eq:#1}}

\long\def\ignorethis#1{}

\newlength\paramargin
\newlength\figmargin
\newlength\subfigmargin
\newlength\secmargin
\newlength\subsecmargin
\newlength\tabmargin
\newlength\eqmargin

\newcolumntype{C}[1]{>{\centering\let\newline\\\arraybackslash\hspace{0pt}}m{#1}}

\newcommand*\colourmark[1]{%
  \expandafter\newcommand\csname #1xmark\endcsname{\textcolor{#1}{\ding{56}}}%
}

\colourmark{blue}
\colourmark{red}

\newcommand*\colourchecksnow[1]{%
  \expandafter\newcommand\csname #1snow\endcsname{\textcolor{#1}{\ding{100}}}%
}

\colourchecksnow{cyan}

\definecolor{darkspringgreen}{rgb}{0, 0.6, 0.3}
\newcommand*\colourcheck[1]{%
  \expandafter\newcommand\csname #1check\endcsname{\textcolor{#1}{\ding{52}}}%
}

\colourcheck{blue}
\colourcheck{cadmiumgreen}

\newcommand*\colourtri[1]{%
  \expandafter\newcommand\csname #1tri\endcsname{\textcolor{#1}{\ding{115}}}%
}

\colourtri{black}

\makeatletter
\@namedef{ver@everyshi.sty}{}
\makeatother
\usepackage{tikzsymbols}
\newcommand*\colourcheckfire[1]{%
  \expandafter\newcommand\csname #1fire\endcsname{\textcolor{#1}{\Fire}}%
}
\colourcheckfire{red}

%% file: 0_Abstract.tex
\begin{abstract}
In this paper, we introduce zero-shot audio-video editing, a novel task that requires transforming original audio-visual content to align with a specified textual prompt without additional model training.
To evaluate this task, we curate a benchmark dataset, \ourdataa, designed explicitly for zero-shot audio-video editing.
\ourdata includes 110 videos, each with a 10-second duration, spanning 11 categories from VGGSound. It offers diverse prompts and scenarios that require precise alignment between auditory and visual elements, enabling robust evaluation.
We identify limitations in existing zero-shot audio and video editing methods, particularly in synchronization and coherence between modalities, which often result in inconsistent outcomes.
To address these challenges, we propose \ourmodela, a zero-shot cross-modal delta denoising framework that leverages audio-video interactions to achieve synchronized and coherent edits.
\ourmodel demonstrates superior results on both \ourdata and the recent OAVE dataset to validate its generalization capabilities.
Results are available at \url{https://genjib.github.io/project_page/AVED/index.html}

\end{abstract}  

%% file: 1_intro.tex
\vspace{\secmargin}
\section{Introduction}\label{sec:intro}
\vspace{\secmargin}
\input{figure/teaser}
Recent advancements in diffusion-based generative models~\cite{ho2020denoising,song2021scorebased,nichol2021improved,dhariwal2021diffusion,moviegen} have demonstrated remarkable progress in image~\cite{ramesh2021dalle,ramesh2022hierarchical,saharia2022photorealistic,rombach2022highresolution,nichol2021glide}, video~\cite{singer2022makeavideo,ho2022imagenvideo,blattmann2023videoldm,hong2023cogvideo,wu2022nuwa}, music~\cite{schneider2023mousai,huang2023noise2music,li2023jen1,lin2025vmas,li2024muvi,tian2025vidmuse}, and audio~\cite{audioldm,audioldm2,huang2023makeanaudio,majumder2024tango2,multifoley,images_that_sound} generation.
While these models deliver impressive quality, the development of adaptable and controllable generative models for real-world applications remains challenging.
This challenge stems from the difficulty of disentangling specific attributes within diffusion models, limiting fine-grained control over generated content.
To address the limitations of controllability in generative models, recent work~\cite{sdedit,avrahami2022blended,kim2022diffusionclip,saharia2022palette,kim2021ilvr,controlnet,instructpix2pix} has focused on enhancing precision and flexibility in content creation. 
These models enable users to create and edit images with fine-grained control, supporting applications like photo editing and personalized content generation. While these approaches have advanced image-based editing, achieving seamless and synchronized edits across both audio and video modalities remains challenging.
However, such capabilities are increasingly crucial for content creators, filmmakers, or digital artists, who require intuitive tools to edit and modify multimedia content efficiently.  
To illustrate the challenges of audio-video editing, consider a scenario in \figref{teaser} where a dog is barking in a room. Suddenly, the dog transforms into a fierce lion with a corresponding roar, enhancing suspense and surprise. 
This scenario requires a model capable of not only transforming the visual appearance of the dog into a lion but also synchronously updating the audio to match the new visual context—tasks that current single-modality editing approaches~\cite{rave,tokenflow,DreamMotion,zs_audio_ddpm,audioeditor,singer2024video,sheynin2024emu} struggle to accomplish.

Efficiently and seamlessly editing real-world audio-video content without substantial computational overhead remains challenging. 
It typically relies on additional datasets~\cite{ho2022imagenvideo,ho2022videodiffusion,singer2022makeavideo,wang2023audit} or finetuning on pretrained text-guided models~\cite{bar2022text2live,kasten2021layered,lee2023shape}.
To reduce these costs, recent zero-shot methods have been proposed for video~\cite{avrahami2022blended,p2p,pnp,fatezero,Text2video-zero,wang2023zero,rave} and audio editing~\cite{zs_audio_ddpm,audioeditor} by leveraging pretrained text-to-image~\cite{rombach2022highresolution,sd3,controlnet} and text-to-audio~\cite{audioldm,audioldm2,stable_audio} diffusion models.
However, existing approaches are limited to single modalities and lack frameworks designed for joint audio-video editing, highlighting a gap in multimodal editing models and benchmarks.

\input{table/datasize}

To mitigate this gap, we introduce the \ourdata dataset, manually curated from VGGSound~\cite{vggsound}.
Unlike video-only datasets in \tabref{size}, \ourdata contains a rich diversity of natural audio-visual events. 
\ourdata consists of 110 distinct 10-second videos across 11 categories, each paired with human-annotated source and target prompts across various categories, including animals, human actions, and environmental sounds.
Some example prompts in \ourdata, such as a dog barking, gun shooting, etc., present a unique challenge that requires precise control over audio-visual changes and synchronization.

To achieve fine-grained and synchronized editing across audio and video, we propose \ourmodela, a zero-shot cross-modal delta denoising framework that jointly edits both modalities while maintaining temporal and structural coherence.  
Unlike existing methods for zero-shot video~\cite{rave,ground-a-video,tokenflow} or audio editing~\cite{zs_audio_ddpm,liu2024medic}, which process audio and video independently, these approaches often result in a misalignment between audio and visual content.
Such limitations occur because existing methods lack a unified approach to aligning audio and video transformations holistically. 
To address this, built upon score distillation~\cite{cds,dds,sds,DreamMotion}, \ourmodel iteratively refines the content by aligning the noise gradients with textual prompts and patch-level audio-visual information.  
At each denoising step, \ourmodel encodes audio and video into a latent space guided by textual prompts and enforces cross-modal consistency through a contrastive loss at the patch level.  
This ensures that edits are coherent and synchronized, maintaining both semantic meaning and temporal consistency across modalities.

We validate \ourmodel on \ourdataa, a benchmark dataset that we manually curated with human-annotated prompts, which consists of a wide variety of natural audio-visual events to evaluate the zero-shot audio-video editing task effectively.
Additionally, we evaluate \ourmodel on the OAVE dataset~\cite{oave}, which is designed for one-shot joint audio-image editing tasks.
The experimental results demonstrate that our cross-modal design outperforms baselines focused solely on either video or audio editing~\cite{rave,ground-a-video,tokenflow,DreamMotion,zs_audio_ddpm}.

%% file: figure/teaser.tex
\begin{figure}[t!]
    \centering
	\includegraphics[width=0.90\linewidth]{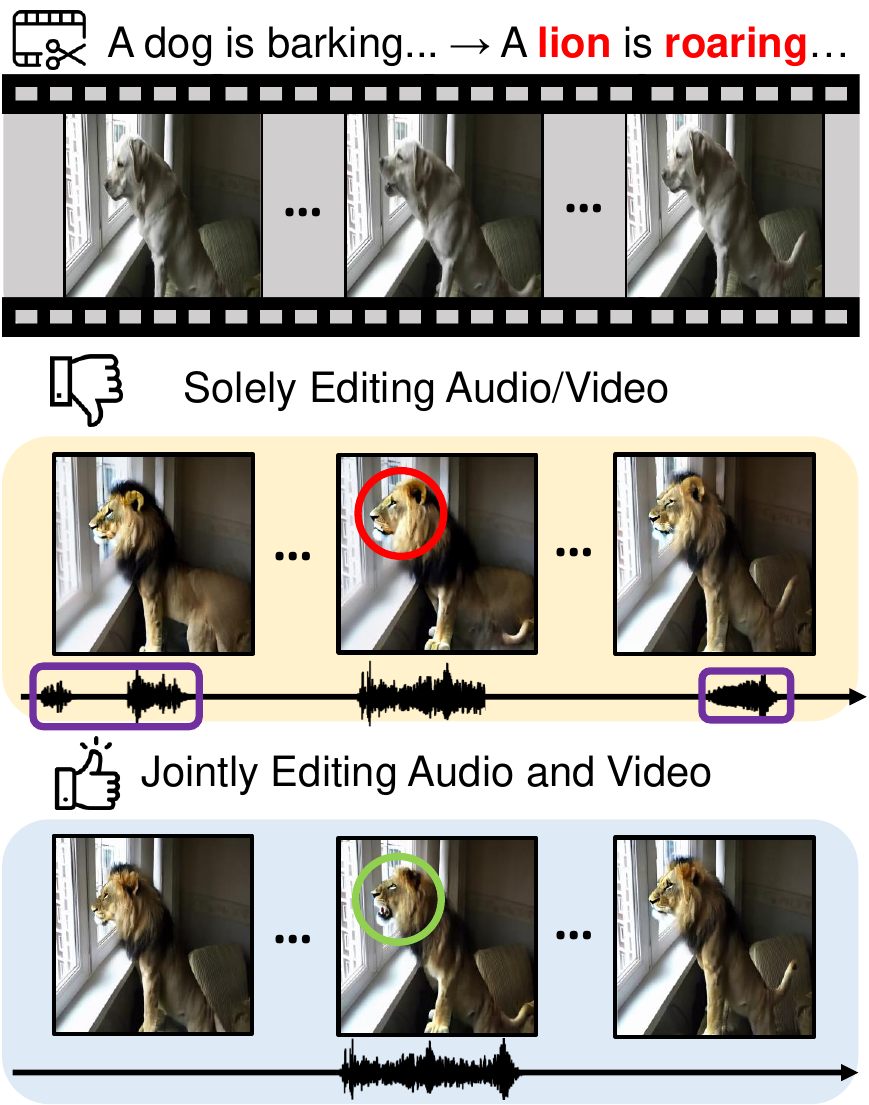}
    \caption{
    \textbf{Key Challenges in Joint Audio-Video Editing.} Existing methods primarily focus on zero-shot text-to-video~\cite{cohen2024slicedit,yang2023rerender_a_video,liu2024video_p2p} or text-to-audio~\cite{zs_audio_ddpm,wang2023audit,audioeditor} editing separately.
    Solely editing only video or only audio often leads to coherence and synchronization issues between two modalities. As highlighted in \textcolor{red}{\textbf{red}} circle, the motion or presence of sounding objects may not align with the corresponding audio.
    Additionally, edited content may exhibit audio artifacts along the temporal dimension (shown in the \textcolor{violet}{\textbf{purple}} squares).
    These factors make the edited results feel less natural and cohesive.
    In contrast, our \ourmodel jointly edits audio and video by leveraging cross-modal information as additional supervision to improve editing quality to alleviate synchronization issues. 
    }
	\label{fig:teaser}
\vspace{\figmargin}
\end{figure}

%% file: table/datasize.tex
\begin{table}[t]
    \centering
    \setlength{\tabcolsep}{6pt} 
    \resizebox{0.49\textwidth}{!}{ 
        \begin{tabular}{lcccc}
            \toprule
            Method & Venue & \# Videos & Modality \\
            \midrule
            DreamMotion~\cite{DreamMotion} & ECCV'24 & 26 & Video \\
            RAVE~\cite{rave} & CVPR'24 & 31 & Video \\
            TokenFlow~\cite{tokenflow} & ICLR'24 & 61 & Video \\
            \ourdataa & N/A & 110 & Video+Audio \\
            \bottomrule
        \end{tabular}
    }
    \caption{
    \textbf{Existing Evaluation Sets in Video-Based Zero-Shot Editing.} 
    Unlike prior video-only benchmarks, \ourdata introduces both video and audio, which is more challenging and enables a comprehensive evaluation of zero-shot audio-video editing.
    }
    \vspace{\tabmargin}
    \label{tab:size}
\end{table}

%% file: 2_related_work.tex
\input{figure/fig_method}

\vspace{\secmargin}
\section{Related Work}
\vspace{\secmargin}
\subsection{Joint Audio-visual Generation}
Jointly generating audio-video content presents a challenging task compared to single-modal generation (\ie video-to-audio~\cite{su2020audeo,pascual2024masked_v2a,chen2024semanticall_v2a,wang2024frieren,diff_foley} or audio-to-video generation~\cite{tpos,zhang2024audio,biner2024sonicdiffusion,lee2022sound}). 
Previous methods have addressed this by introducing novel audio and video tokenizers for effective autoregressive multi-modal generation~\cite{wang2024avdit,kim2024versatile_dit,ishii2024simple,mmdiffusion,codi,codi2}, training diffusion models on paired audio-video data~\cite{wang2024avdit,kim2024versatile_dit,mmdiffusion}, or combining multiple single-modal diffusion models with cross-modal alignment techniques~\cite{codi,codi2,ishii2024simple}. 
While these methods have shown promising results in modeling audio-video data, achieving precise control over the generated content remains challenging, particularly when editing audio-video inputs to align with specific prompts.

A recent study~\cite{oave} investigates joint audio-image editing by finetuning pretrained text-to-image and text-to-audio models on a small set of text-audio-image triplets (\ie one-shot setting) for each editing sample.
Unlike audio-image editing~\cite{oave} or audio-video generation~\cite{wang2024avdit,kim2024versatile_dit,mmdiffusion}, we explore a new task, zero-shot audio-video editing, which is inherently more challenging than image-only tasks due to the rich temporal information across modalities (\eg temporal consistency and synchronization issues). 
Besides, the proposed \ourmodel eliminates the need for additional finetuning of diffusion models, making it more computationally efficient than the finetuned model~\cite{oave}.
\subsection{Audio/Video Editing using Diffusion Models}
\vspace{\secmargin}
Recent works~\cite{avrahami2022blended,p2p,pnp,fatezero,Text2video-zero,wang2023zero,rave,zs_audio_ddpm,cds,DreamMotion,sds,dds,liu2024video_p2p,zhang2023controlvideo} leverage pretrained text-to-image~\cite{rombach2022highresolution,sd3,controlnet} and text-to-audio~\cite{audioldm,audioldm2,huang2023makeanaudio} to achieve content editing based on text prompts. 
Among various of these techniques~\cite{avrahami2022blended,dds,sdedit}, inverting source embeddings into noise vectors, such as DDIM and DDPM inversion, achieves efficient zero-shot editing for visual or audio data based on text prompts.
Text-based audio editing models~\cite{zs_audio_ddpm,audioeditor} leverage DDIM/DDPM inversion and pretrained diffusion models~\cite{audioldm,audioldm2,stable_audio} to achieve zero-shot editing without model finetuning or additional training datasets~\cite{wang2023audit}.
In video editing, recent works~\cite{rave,tokenflow,liu2024video_p2p,zhang2023controlvideo,DreamMotion,fatezero,cohen2024slicedit,wu2023tune_a_video,li2024vidtome,fan2024videoshop,yu2025veggie,yoon2024raccoon} have extended diffusion-based image editing methods~\cite{cds,dds,sdedit,p2p} to the video domain by improving the temporal consistency of edits in both zero-shot and few-shot settings.
In particular, these works~\cite{fatezero,cohen2024slicedit,li2024vidtome} inflate spatial self-attention layers to spatiotemporal layers to enhance video frame coherence by leveraging the effectiveness of inversion techniques.
Another branch of work~\cite{tokenflow,rave,yang2023rerender_a_video} seeks to maintain the structure of source videos by incorporating structural cues, such as optical flow~\cite{yang2023rerender_a_video,tokenflow}, or depth maps~\cite{rave,ground-a-video}.

Beyond inversion methods, Score Distillation Sampling (SDS)~\cite{sds} refine images and videos by calculating gradients from pretrained diffusion models based on target prompts. 
While SDS enables selective editing of prompt-relevant areas, it can suffer from over-smoothing and over-saturation.
To mitigate this, Delta Denoising Score (DDS)~\cite{dds} introduces a reference branch with source and target prompts for SDS and demonstrates promising editing results on different tasks~\cite {cds,DreamMotion}.
However, these approaches~\cite{rave,tokenflow,cds,dds,zs_audio_ddpm,audioeditor}, solely editing audio and video, still suffer from the synchronization and coherence issues across two modalities.
To address these limitations, we propose \ourmodel with a cross-modal delta denoising scheme that jointly leverages audio and video information.
This scheme provides cross-modal supervision to yield synchronized and high-fidelity edits in both audio and video.

%% file: figure/fig_method.tex
\begin{figure*}[t!]
    \centering
	\includegraphics[width=0.9\linewidth]{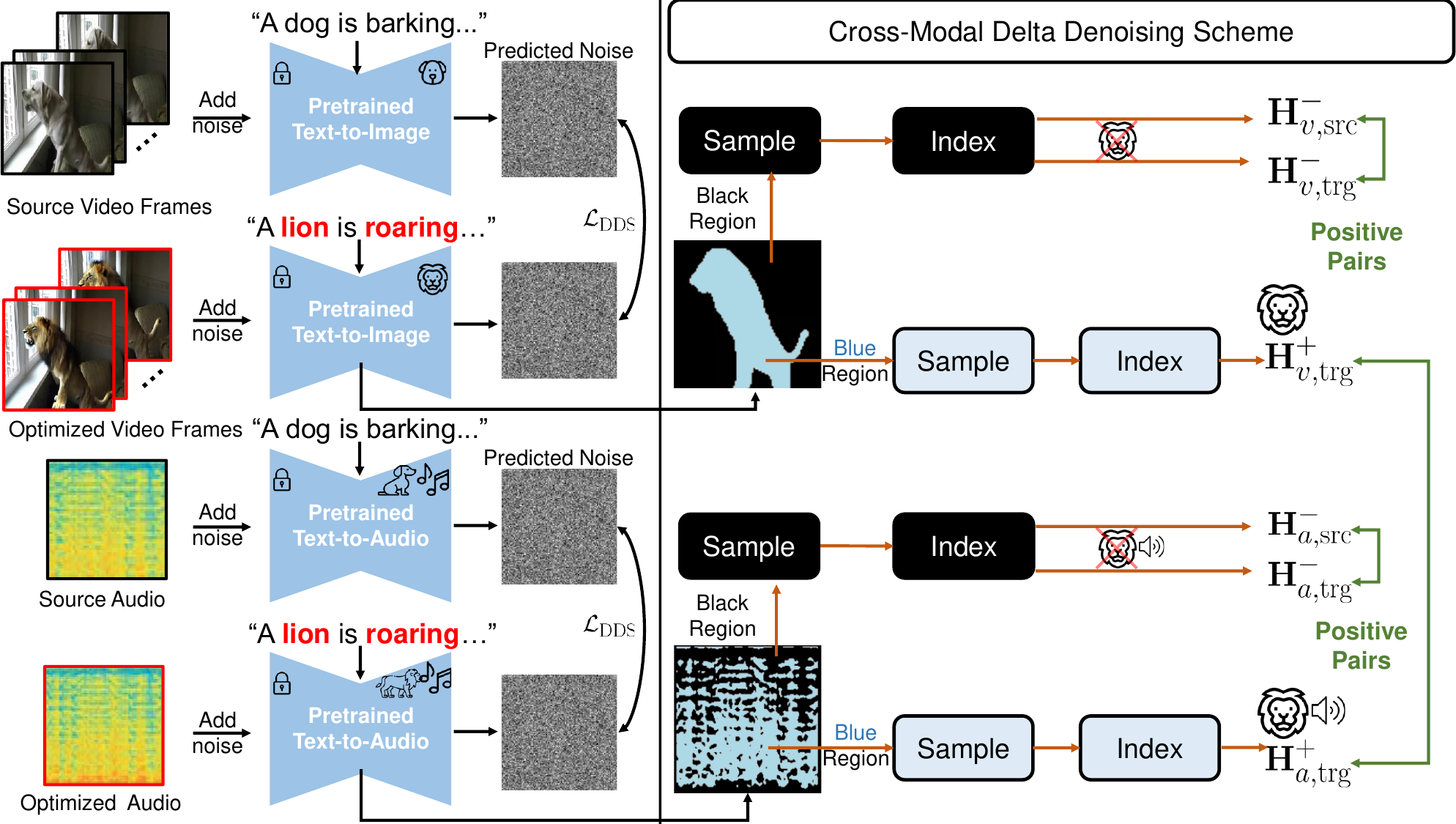}
    \caption{\textbf{Our \ourmodel Framework.} \ourmodel performs zero-shot audio-video editing by employing a cross-modal delta denoising score scheme to edit audio and video based on target prompts jointly.
    During the denoising process, relevance scores are computed between audio/image regions and target textual prompts within the cross-attention module from the diffusion model.
    These scores identify prompt-relevant regions (\ie \textcolor{cyan}{\textbf{blue}} areas) and irrelevant patches, allowing selective editing of specific regions while preserving unaltered content.
    Using this region information (obtained by randomly sampling patch indices), we define positive pairs as unaltered content consistent in both the source and target branches and regions requiring edits across audio and video modalities. All other pairs are treated as negative pairs.
    This design enables synchronized, high-fidelity edits aligned with target prompts, maintaining coherence across audio and video.
    }
    \vspace{\figmargin}
	\label{fig:method}
\end{figure*}

%% file: 3_Method.tex
\vspace{\secmargin}
\section{Method}
\vspace{\secmargin}
\label{sec:method}
This section defines the notations used for zero-shot audio-video editing and provides an overview of the Delta Denoising Score (DDS)~\cite{dds}. 
We then introduce the cross-modal denoising scheme in \ourmodela, specifically designed to facilitate zero-shot joint audio-video editing. 
Our approach leverages cross-modal supervision to improve the coherence and quality of audio-visual edits.

\subsection{Preliminaries}
\vspace{\subsecmargin}
\paragraph{Video, Audio, and Text Inputs.}  
Let $\mathbf{z}^{v} \in \mathbb{R}^{F \times d_v \times W_v \times H_v}$ and $\mathbf{z}^{a} \in \mathbb{R}^{d_a \times W_a \times H_a}$ represent the latent features of video and audio, respectively, encoded by a pretrained VAE~\cite{vae}, where $F$ denotes the number of video frames.  
Here, $W_v$ and $H_v$ represent the width and height of video frames, while $W_a$ and $H_a$ denote the dimensions of the audio spectrogram.  
The channel dimensions for video and audio embeddings are given by $d_v$ and $d_a$, respectively.  
Text prompts, denoted as $y_{src}$ and $y_{trg}$, specify the source (optional) and target descriptions of the audio-video content.  
Following RAVE~\cite{rave}, we shuffle the temporal order of video frames and process them as a spatial grid, allowing a pretrained text-to-image diffusion model to perform video-level editing.  
We represent the video latent in a grid format of $n_g \times n_g$ as $\mathbf{z}^{g} \in \mathbb{R}^{M \times d_v \times W_g \times H_g}$, where $M ={F}/{n_g \times n_g}$, $W_g = {W_v}/{n_g}$, and $H_g = {H_v}/{n_g}$.

\noindent\textbf{Delta Denoising Score (DDS).} The Delta Denoising Score (DDS)~\cite{dds} extends the Score Distillation Sampling (SDS)~\cite{sds} by introducing a structured comparison between two distinct textual inputs and the same image inputs, referred to as the source and target branches, for precise editing.
Here, the source branch represents the original, unedited content, while the target branch aligns with the desired edits specified by the target prompt. 
This approach enables more precise control over modifications by refining the difference between these branches.
Specifically, the gradient computation in DDS starts with noise prediction in a text-conditioned diffusion model using classifier-free guidance (CFG)~\cite{ho2022classifier}, which is formulated as follows:
\begin{equation}
\epsilon_{\phi}^{\omega}(\mathbf{z}_{t}, y, t) = (1+{\omega}) \epsilon_{\phi}(\mathbf{z}_{t}, y, t) - \omega \epsilon_{\phi}(\mathbf{z}_{t}, \emptyset, t),
\label{eq:CFG}
\end{equation}
where $\omega$ is the guidance parameter, $\epsilon_{\phi}$ denotes the noise prediction network, and $\emptyset$ is the null-text prompt.
Here, $y$ represents either the source or target text prompt, and $\mathbf{z}_t$ can be an audio or image latent at a timestamp $t \sim \mathcal{U}(0, 1)$ sampled from a uniform distribution.
DDS computes gradients based on both target and reference inputs, defined by the following objective:
\begin{equation}
\mathcal{L}_{\mathrm{DDS}}(\theta; y_{trg}) = \|\epsilon_{\phi}^{\omega}(\mathbf{z}_t(\theta), y_{trg}, t) - \epsilon_{\phi}^{\omega}(\mathbf{z}_t, y_{src}, t)\|^2,
\label{eq:lossDDS}
\end{equation}
where $\mathbf{z}_t(\theta)$ represents the target latent, parameterized by $\theta$.
$\mathbf{z}_t(\theta)$ is iteratively updated in the direction of the gradient $\nabla_{\theta}\mathcal{L}_{\mathrm{DDS}}$ to adjust toward the target prompt precisely.

\subsection{Cross-Modal Delta Denoising Scheme}  
\vspace{\subsecmargin}
We introduce a cross-modal delta denoising scheme to improve the quality and coherence of audio-video editing by leveraging interactions between modalities during each DDS process. Unlike single-modality approaches, our method incorporates complementary information from both audio and video, ensuring synchronized and contextually consistent edits.   
We first identify \textit{\textbf{prompt-relevant regions}} in audio and video to achieve this.  
This is done by computing similarity scores within the cross-attention modules of the diffusion layers.  
These scores indicate how well different patches in the audio and video streams align with the prompts.  
By applying thresholding, we classify relevant patches (aligned with the prompt) and irrelevant patches (background or unchanged content).  
Once identified, we sample pairs from the source (\ie $\epsilon_{\phi}^{\omega}(\mathbf{z}_t, y_{src}, t)$) and target (\ie $\epsilon_{\phi}^{\omega}(\mathbf{z}_t(\theta), y_{trg}, t)$) prompts.  
These sampled pairs serve as the foundation for a contrastive loss, which enforces alignment between corresponding regions across modalities, improving synchronization and overall coherence.

\noindent\textbf{Prompt-Relevant Patches.}  To identify regions in the audio and video relevant to prompts, we leverage intermediate representations in the cross-modal attention layers of pretrained diffusion models.
By computing the similarity between audio/video features (queries) and textual prompt features (keys), we highlight areas aligned with prompts.  

Let $\mathbf{Q}_{a} \in \mathbb{R}^{n_{q} \times d}$ and $\mathbf{Q}_{v} \in \mathbb{R}^{M \times n_{q} \times d}$ be the audio and video query features, while $\mathbf{K}_{a} \in \mathbb{R}^{n_{k} \times d}$ and $\mathbf{K}_{v} \in \mathbb{R}^{M \times n_{k} \times d}$ represent their corresponding key features derived from the target prompt.  
The similarity scores $\mathbf{S}$ are computed as:  
\begin{equation}
\mathbf{S}_{i}^{a} = \max_{j} \left( \mathbf{Q}_{a} \mathbf{K}_{a}^{\top} \right)_{i,j}, \quad
\mathbf{S}_{i}^{v} = \max_{j} \left( \mathbf{Q}_{v} \mathbf{K}_{v}^{\top} \right)_{i,j}.
\end{equation}
Here, $\mathbf{S}^a \in \mathbb{R}^{n_{q}}$ and $\mathbf{S}^v \in \mathbb{R}^{M \times n_{q}}$ represent the prompt relevance scores for audio and video patches.  
To focus on the most relevant regions, we apply max pooling across the text dimension \( j \).  
We then normalize the scores using min-max normalization to ensure values range between 0 and 1 for consistency.  
The final normalized scores, $\tilde{\mathbf{S}}^{a}_{\text{trg}}$ and $\tilde{\mathbf{S}}^{v}_{\text{trg}}$, will be thresholded to distinguish relevant and irrelevant patches for further processing.

\noindent\textbf{Contrastive Loss for Denoising.} 
Unlike typical contrastive loss, which typically maximizes the similarity between features in a batch (e.g., different audio-video samples), we leverage contrastive loss on relevant and irrelevant patches within the same audio/video instance, which gradually transforms over timesteps by the DDS process, to improve coherence in audio-video editing.  
For example, when editing a video where a dog transforms into a lion, we aim to align audio and video regions such as the lion’s roar and fur as positive pairs, while ensuring that the background or unrelated objects remain unchanged. 
To preserve context, irrelevant patches with the same spatial location in both the source and target branches of each modality (e.g., background) are also treated as positive pairs. 
All other combinations are considered negative pairs to achieve precise and contextually consistent edits.

To achieve this, we define $\mathcal{I}^{+}_{a}$ and $\mathcal{I}^{+}_{v}$ as indices for patches in the target audio and video branches aligned with a textual prompt, where $\tilde{\mathbf{S}}^{a}_{\text{trg}} > \tau_{a}$ or $\tilde{\mathbf{S}}^{v}_{\text{trg}} > \tau_{v}$ with given threshold $\tau_{a}$ and $\tau_{v}$. 
These relevant patches correspond to areas that need editing, such as the sound and appearance of the lion when it transforms from a dog.
Similarly, $\mathcal{I}^{-}_{a}$ and $\mathcal{I}^{-}_{v}$ represent indices for patches in the target branch that are not related to the prompt and do not require modification, such as background sounds or static visual elements where $\tilde{\mathbf{S}}^{a}_{\text{trg}} < \tau_{a}$ or $\tilde{\mathbf{S}}^{v}_{\text{trg}} < \tau_{v}$.
We then extract audio embeddings $\mathbf{h}_{a} \in \mathbb{R}^{n_{q}^{a} \times d_a}$ and video embeddings $\mathbf{h}_{v} \in \mathbb{R}^{M \times n_{q}^{v} \times d_v}$ from the hidden states of cross-modal attention layers to capture rich information for precise editing.
To achieve more diverse views and robust editing, we randomly sample embeddings for relevant regions from the target branch:
\begin{equation}
\mathbf{H}_{a, \text{trg}}^{+} = \{ \mathbf{h}_{a, \mathrm{trg}, i} \mid i \in \mathcal{I}^{+}_{a} \}, \quad \mathbf{H}_{v, \text{trg}}^{+} = \{ \mathbf{h}_{v, \mathrm{trg}, i} \mid i \in \mathcal{I}^{+}_{v} \}.
\end{equation}
Here, $\mathbf{H}_{a}^{+}$ and $\mathbf{H}_{v}^{+}$ represent sets of relevant audio and video embeddings sampled from the target branch.
Similarly, for irrelevant regions, we sample embeddings directly from both the source and target branches:
\begin{equation}
\begin{aligned}
\mathbf{H}_{a, \text{src}}^{-} &= \{ \mathbf{h}_{a, \mathrm{src}, i} \mid i \in \mathcal{I}^{-}_{a} \}, \quad \mathbf{H}_{a, \text{trg}}^{-} = \{ \mathbf{h}_{a, \mathrm{trg}, i} \mid i \in \mathcal{I}^{-}_{a} \}, \\
\mathbf{H}_{v, \text{src}}^{-} &= \{ \mathbf{h}_{v, \mathrm{src}, i} \mid i \in \mathcal{I}^{-}_{v} \}, \quad \mathbf{H}_{v, \text{trg}}^{-} = \{  \mathbf{h}_{v, \mathrm{trg}, i} \mid i \in \mathcal{I}^{-}_{v} \}.
\end{aligned}
\end{equation}
Here, $\mathbf{H}_{a}^{-}$ and $\mathbf{H}_{v}^{-}$ represent sets of irrelevant audio and video embeddings sampled from both branches.
We then apply a contrastive loss that encourages high similarity between positive pairs, including relevant patches and irrelevant patches matched by the same index across source and target branches within the same modality while discouraging similarity with unrelated negative pairs.
The standard contrastive loss is described as follows:
\begin{equation}
\begin{aligned}
\label{eq:contrastive}
\mathcal{L}_{c}(\mathbf{F}_x, \mathbf{F}_y) = -\frac{1}{N} \sum_{i=1}^N \log \frac{\exp \left( \text{sim}(\mathbf{F}_{x}^{i}, \mathbf{F}_{y}^{i}) / \alpha \right)}{\sum_{j=1}^{N} \exp \left( \text{sim}(\mathbf{F}_{x}^{i}, \mathbf{F}_{y}^{j}) / \alpha \right)},
\end{aligned}
\end{equation}
where $\text{sim}(\cdot, \cdot)$ denotes cosine similarity, $N$ is the mini-batch size, and $\alpha$ is a temperature parameter.
Our final cross-modal contrastive loss combines alignments as follows:
\begin{equation}
\begin{aligned}
\mathcal{L}_{\text{cmds}} =  \frac{1}{M}& \sum_{i=1}^{M} \Big(\mathcal{L}_{c} \big( [ \mathbf{H}_{a,\text{trg}}^{+}, \mathbf{H}_{v, \text{src}}^{-}[i] ], [ \mathbf{H}_{v,\text{trg}}^{+}[i], \mathbf{H}_{a, \text{src}}^{-} ] \big) \\
&+ \mathcal{L}_{c} \big( [ \mathbf{H}_{v,\text{trg}}^{+}[i], \mathbf{H}_{a, \text{src}}^{-} ], [ \mathbf{H}_{a,\text{trg}}^{+}, \mathbf{H}_{v, \text{src}}^{-}[i] ] \big) \Big).
\end{aligned}
\label{eq:losscmds}
\end{equation}
The proposed $\mathcal{L}_{\text{cmds}}$ not only leverages cross-modal information to improve editing quality, but also preserves unedited content by considering unrelated patches across branches in the same modality. 
For instance, when converting a barking dog into a roaring lion, the generated lion’s roar ($\mathbf{H}_{a,\text{trg}}^{+}$) should be distinct from non-dog sounds in the source audio ($\mathbf{H}_{a,\text{src}}^{-}$), and the lion’s appearance ($\mathbf{H}_{v,\text{trg}}^{+}$) should differ from regions in the original video that lack lion-related features ($\mathbf{H}_{v,\text{src}}^{-}$).  
If $d_v$ and $d_a$ differ in some layers, we simply interpolate to match the dimensions. 
The final objective integrates this contrastive loss with DDS, formulated as $\mathcal{L}_{\text{cmds}} + \mathcal{L}_{\mathrm{DDS}}(\theta; y_{\text{trg}})$.\footnote{Implementation details are provided in the supplementary material.}

%% file: 4_result.tex
\input{table/sota}

\vspace{\secmargin}
\section{Experimental Setup}
\vspace{\secmargin}
\label{sec:exp_setup}
\subsection{Downstream Datasets}
\vspace{\subsecmargin}
\begin{compactitem}
    \item \textbf{\ourdataa}\footnote{Full annotations are included in the supplementary material.}, our newly curated dataset, consists of $110$ 10-second videos in $11$ distinct categories from VGGSound~\cite{vggsound}, covering diverse scenes such as animals, objects, and environmental sounds.
    Each video is paired with annotated source and target prompts that specify audio-visual events and object categories. 
    This dataset provides a comprehensive benchmark for evaluating zero-shot audio-video editing capabilities.
    \item \textbf{OAVE}~\cite{oave} contains $44$ categories, each with 10 images from a clip and separate audio and visual annotations.  
    It includes $25$ prompt templates to modify either the sounding object or the environmental context for evaluating editing performance.
\end{compactitem}

\subsection{Evaluation Metrics}
\vspace{\subsecmargin}
\label{sec:evaluation}
Following previous work~\cite{rave,cds,dds,tokenflow,zs_audio_ddpm,sdedit}, we evaluate our generated audio-video samples using several metrics:
\textbf{CLIP-F} evaluates the consistency of edited frames by calculating the similarity of frame-based CLIP embeddings~\cite{clip}.
\textbf{CLIP-T} evaluates the alignment between the target prompt and the edited video by computing the CLIP similarity~\cite{clip} between each video frame and the prompt, then averaging these similarity scores across all frames.
\textbf{DINO} emphasizes the preservation of the overall structure between the source and target frames by computing cosine similarity with self-supervised DINO-ViT embeddings~\cite{dino}.
\textbf{Obj} utilizes Grounding-DINO~\cite{gdino} to detect and assess the presence and likelihood of target objects specified in the prompt, evaluating how accurately these objects are generated.
\textbf{CLAP} measures the cosine similarity between audio and target prompt using the CLAP model~\cite{clap}, indicating the fidelity of the edited sound.
\textbf{LPAPS} evaluates perceptual distance in source and target audio in CLAP feature space~\cite{clap} to provide an assessment of faithfulness/consistency between edited and source sound.
\textbf{AV-Align}~\cite{yariv2024diverse} metric examines alignment between audio cues and visual changes, low-level coherence in audio-visual transitions.
\textbf{IB} leverages ImageBind~\cite{girdhar2023imagebind} embeddings to assess audio-visual similarity to evaluate high-level audio-visual coherence.

\input{figure/human_bar}
\subsection{Human Evaluation}
\vspace{\subsecmargin}
\label{sec:human_setting}
We conduct a human evaluation by asking subjects to select their preferred edited audio-video samples according to their alignment with the target prompt.
Specifically, given a source (unedited) video and a pair of edited audio-video samples, human raters are asked to select their preferred sample based on the following question: \textit{Which video do you think has the better editing quality overall?}
For each question, subjects can choose one of the two methods or a third option, ``Cannot tell." 
Each subject evaluates five randomly selected video pairs, with one sample always from \ourmodela.  
To prevent bias, the methods remain unknown to the raters.  
We compare our \ourmodel method against competing approaches~\cite{zhang2023controlvideo,tokenflow,rave}. 
Results are reported as the average human preference rate for each method.  
Our human study is conducted with $300$ subjects on Amazon Mechanical Turk.

\input{table/sota_oave}

\subsection{Baselines}
\vspace{\subsecmargin}
We compare our model to recent baselines in zero-shot audio or video editing, evaluating each modality independently due to the lack of joint audio-visual approaches.
For video editing, we assess:
(i) RAVE~\cite{rave}, which uses pretrained text-to-image diffusion models and noise-shuffling for temporally consistent edits, 
(ii) TokenFlow~\cite{tokenflow}, which maintains feature consistency through inter-frame correspondences for high-quality edits without extra training, and 
(iii) ControlVideo~\cite{zhang2023controlvideo}, which adapts ControlNet~\cite{controlnet} for training-free text-to-video generation using depth maps and human poses.
For audio editing, we evaluate:
(i) ZEUS~\cite{zs_audio_ddpm}, which edits via DDPM inversion and reversion to edit sound, and 
(ii) SDEdit~\cite{sdedit}, which blends denoising processes with initial noise to edit audio outputs.
We implement a sequential baseline, RAVE~\cite{rave} $\rightarrow$ Diff-Foley~\cite{diff_foley}, where RAVE edits video first, followed by Diff-Foley for audio generation. We also implement DDS~\cite{dds}, which applies the same methodology separately to audio and video using image grids and shuffling.  
This evaluation allows us to compare the performance of \ourmodel with leading methods in both domains.

\vspace{\secmargin}
\section{Results and Analysis}
\vspace{\secmargin}
\label{sec:results}
%

%
%
%

\input{table/abs_single}
\subsection{Comparison with the State-of-the-Art}
\vspace{\subsecmargin}
In \tabref{sota}, we present a detailed comparison of \ourmodel with the leading zero-shot video and audio editing models~\cite{zhang2023controlvideo,tokenflow,rave,dds,zs_audio_ddpm,sdedit} and the sequential baseline.
We evaluate performance on video-only, audio-only, and joint audio-video metrics.
For \textbf{video-only metrics}, \ourmodel consistently outperforms baseline models, achieving a notable increase in DINO scores (e.g., \textbf{0.956} vs. \textbf{0.921} for DDS~\cite{dds}), which highlights \ourmodela’s strength in preserving visual coherence and structural fidelity between source and edited video frames. 
Furthermore, \ourmodel achieves the highest CLIP-F, CLIP-T, and Obj. scores (\textbf{0.903}, \textbf{0.260}, and \textbf{0.180}, respectively), demonstrating its ability to maintain frame consistency and align edits precisely with the target prompt. %

For \textbf{audio-only metrics}, \ourmodel achieves the highest CLAP score (\textbf{0.226}) and the lowest LPAPS score (\textbf{5.55}).
Compared to DDS~\cite{dds}, ZEUS~\cite{zs_audio_ddpm}, and SDEdit~\cite{sdedit}, \ourmodel shows a significant improvement in LPAPS scores (\ie  \textbf{5.55} vs. \textbf{5.93}, \textbf{6.41}, and \textbf{6.93})  indicating superior perceptual consistency between the source and edited audio.
We note that \ourmodel also outperforms the sequential baseline, achieving higher CLAP (\textbf{0.226} vs. \textbf{0.191}) and lower LPAPS (\textbf{5.55} vs. \textbf{7.33}), demonstrating that the sequential approach leads to degraded audio quality due to potential imperfect results from edited videos.

For \textbf{joint audio-video metrics}, \ourmodel demonstrates a remarkable improvement, achieving over a \textbf{20\%} relative improvement in AV-Align scores compared to baselines. 
This result validates \ourmodel's ability to synchronize visual and audio edits accurately, aligning actions and sounds seamlessly.
Furthermore, in the ImageBind (IB) score, \ourmodel also presents the highest score, suggesting a better high-level audio-video alignment. 
Compared to the sequential baseline, \ourmodel also achieves a higher IB (\textbf{0.23} vs. \textbf{0.16}) and AV-Align (\textbf{0.42} vs. \textbf{0.35}), indicating that the sequential approach accumulates misalignment errors from imperfect video edits and degraded audio quality. 

In \tabref{sota_oave}, we evaluate \ourmodel on the OAVE dataset, a recent benchmark for one-shot audio-image editing.  
Since this task primarily focuses on audio-image editing, metrics such as CLIP-F and AV-Align are less applicable.  
The results show that \ourmodel outperforms baselines across key metrics, achieving the highest DINO score (\textbf{0.959} vs. \textbf{0.930}) and the lowest LPAPS score (\textbf{5.15} vs. \textbf{5.58}), demonstrating its ability to maintain coherence and structural fidelity within audio-visual content.  
Furthermore, \ourmodel achieves a notable increase in the IB score (\textbf{0.32} vs. \textbf{0.29}), indicating a stronger alignment of audio-visual semantics.

The results in both \ourdata and OAVE~\cite{oave} validate the effectiveness of our cross-modal delta denoising scheme in preserving structural consistency (DINO, LPAPS), maintaining perceptual fidelity (CLIP-T, CLAP), and ensuring aligned audio-visual content (AV-Align, IB).  
These findings underline the necessity of joint audio-video editing, as sequential and single-modality approaches introduce misalignment and degrade perceptual quality.

\textbf{Human Evaluation.} \figref{human} presents the results of our human study to assess the overall quality of the edited audio video based on alignment with the target prompt.  
We report human preference rates, which indicate the percentage of raters who preferred \ourmodel over each baseline method.  
Each comparison is conducted between \ourmodel and the baseline methods, including ControlVideo~\cite{zhang2023controlvideo}, TokenFlow~\cite{tokenflow}, and RAVE~\cite{rave}.  
As shown in \figref{human}, approximately \textbf{75\%} of participants preferred \ourmodel over ControlVideo, and over \textbf{60\%} favored \ourmodel compared to TokenFlow and RAVE, highlighting \ourmodela's superior editing quality.

%

%
%

\input{table/abs_select}
\input{figure/fig_sota}
\subsection{Ablation Studies}
\vspace{\subsecmargin}
\paragraph{Impact of Cross-Modal vs. Single-Modal Delta Denoising Scheme.} In \tabref{abs_single}, we analyze the impact of our proposed cross-modal delta denoising scheme compared to a single-modal denoising approach on \ourdata.
The single-modal denoising approaches (\ie "Audio-Only" and "Video-Only" ) indicate that only one modality's features (either audio or video) are used in \eqnref{losscmds}.
Starting with the baseline (\ie DDS~\cite{dds} baseline), we see improvements using single-modal schemes.

In the audio-only denoising scheme, we observe an LPAPS reduction from \textbf{5.93} (Baseline) to \textbf{5.60}, which demonstrates greater consistency between edited and source sounds, resulting in fewer artifacts and a notable improvement in audio-visual alignment as shown by the AV-Align metric increase from \textbf{0.33} to \textbf{0.36}.
A similar trend is observed in the video-only denoising scheme, where the DINO score increases from \textbf{0.921} to \textbf{0.937}, indicating enhanced visual structure preservation of the source content.
Significant improvements are observed when both audio and visual information are integrated within the cross-modal denoising scheme. 
Compared to the baseline, the cross-modal approach increases the DINO score from \textbf{0.921} to \textbf{0.956} and reduces LPAPS from \textbf{5.93} to \textbf{5.55}, thus significantly boosting the AV-Align metric from \textbf{0.33} to \textbf{0.42}. 
These results show the effectiveness of cross-modal design in leading synchronized and coherent audio-video edits.

\begin{figure*}[t] 
    \centering
    \resizebox{\textwidth}{!}{%
        \includegraphics[width=0.9\textwidth]{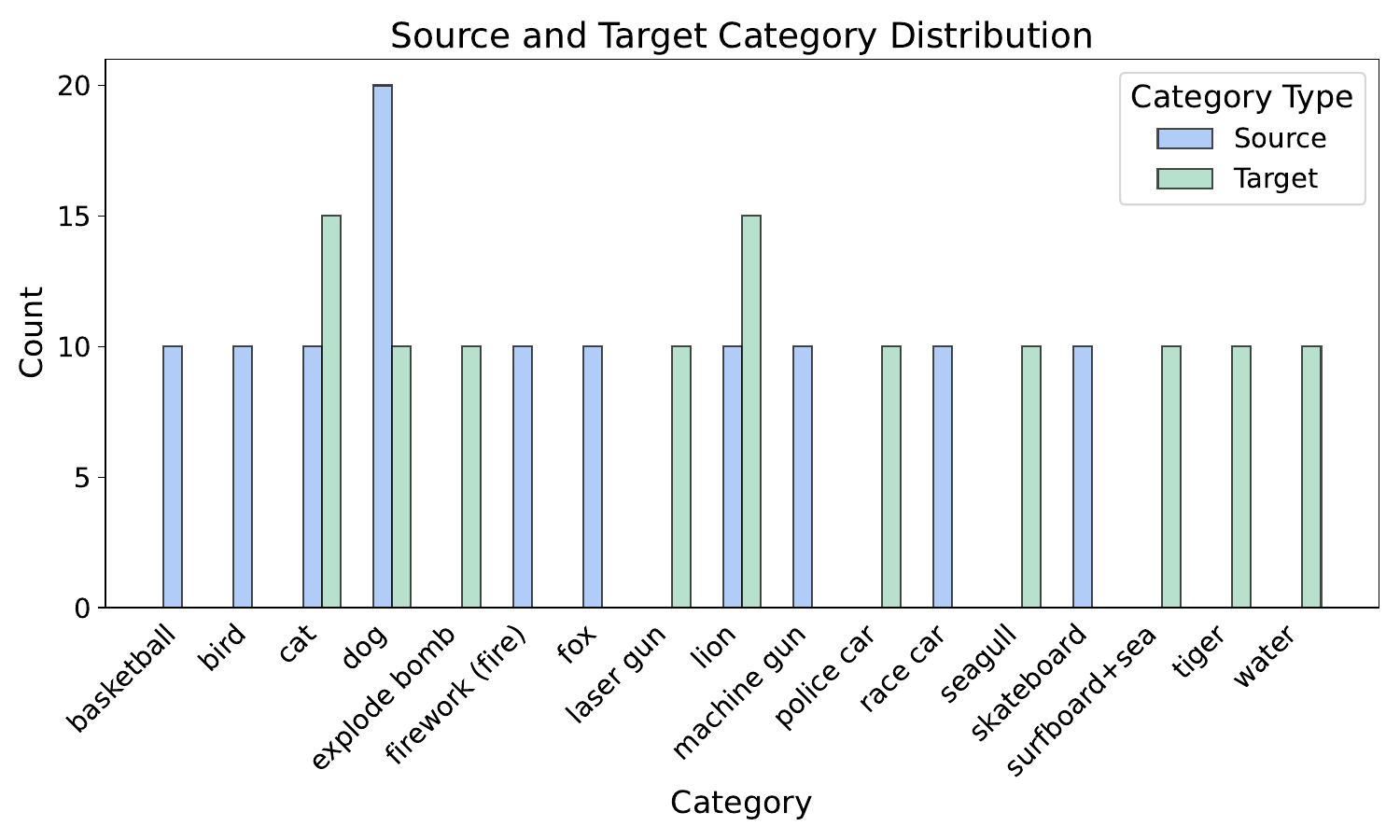}
    }
    \caption{
        \textbf{Category Distribution of \ourdataa.} 
        We present the source and target category distribution of the \ourdata dataset. The source categories represent the initial categories, while the target categories indicate their edited categories.
        This distribution highlights \ourdata's capability to effectively evaluate a variety of audio-video editing.
    }
    \label{fig:data_src_trg}
\end{figure*}

\noindent\textbf{Impact of Selecting Positive and Negative Pairs.}  
Next, we study patch selection strategies in \ourmodel without prompt-relevant information in \tabref{select}.  
The "Random A/V" configuration, which samples positive pairs within the same modality across source and target branches, slightly improves the baseline (\textbf{0.930} DINO, \textbf{5.83} LPAPS, \textbf{0.35} AV-Align).  
Since most patches are unrelated to the editing patches, even random selection provides some advantages.
Instead, the "Random A+V" configuration, assuming all intra-branch audio-video pairs are positive, degrades performance (\textbf{0.902} DINO, \textbf{6.32} LPAPS, \textbf{0.28} AV-Align) since the objects or sounds we aim to edit are usually limited to specific regions. 
Our prompt-relevant selection yields the best results (\textbf{0.956} DINO, \textbf{5.55} LPAPS, \textbf{0.42} AV-Align), confirming its importance for audio-video editing.

\subsection{Qualitative Results}
\vspace{\subsecmargin}
\figref{sota} presents qualitative results of zero-shot audio-video editing for a "Cat" to "Dog" transition.  
\ourmodel is compared with video editing methods, including ControlVideo~\cite{zhang2023controlvideo}, TokenFlow~\cite{tokenflow}, RAVE~\cite{rave}, and the audio model ZEUS~\cite{zs_audio_ddpm}.
The \textcolor{green}{\textbf{green}} circles highlight well-aligned motion in the video frames to demonstrate that \ourmodel accurately transforms the visual appearance and motion, potentially producing sound.
Besides, black rectangles emphasize precise audio matching, demonstrating temporal consistency with the source sound.  
In contrast, competing models exhibit misalignment, as indicated by blue rectangles showing random audio artifacts that disrupt synchronization.  
Overall, this qualitative comparison highlights the effectiveness of \ourmodel’s cross-modal delta denoising scheme in achieving synchronized audio-video edits.

%% file: table/sota.tex
\begin{table*}[t]
    \centering
    \resizebox{0.99\linewidth}{!}{
    \begin{tabular}{c c|cccc|cc|cc }
        \toprule
        \multicolumn{2}{c}{} & \multicolumn{4}{c}{\footnotesize{Video-Only}} & \multicolumn{2}{c}{\footnotesize{Audio-Only}} & \multicolumn{2}{c}{\footnotesize{Joint AV}} \\
        \cmidrule{3-10}
        \textbf{Video Model} & \textbf{Audio Model} & \textbf{CLIP-F}$\uparrow$ & \textbf{CLIP-T}$\uparrow$ & \textbf{Obj.}$\uparrow$ & \textbf{DINO}$\uparrow$ & \textbf{CLAP}$\uparrow$ & \textbf{LPAPS}$\downarrow$ & \textbf{IB.}$\uparrow$ & \textbf{AV-Align}$\uparrow$ \\
        \midrule
        ControlVideo~\cite{zhang2023controlvideo} & SDEdit~\cite{sdedit}        & 0.883 & 0.255 & 0.176 & 0.892  & 0.190 & 6.93 & 0.21 & 0.29 \\
        TokenFlow~\cite{tokenflow}      &  SDEdit~\cite{sdedit}                 & 0.876 & 0.252 & 0.173 & 0.924  & 0.190 & 6.93 & 0.19 & 0.26 \\
        RAVE~\cite{rave}                &  SDEdit~\cite{sdedit}                 & 0.885 & 0.251 & 0.170 & 0.881  & 0.190 & 6.93 & 0.18 & 0.29 \\
        \midrule
        ControlVideo~\cite{zhang2023controlvideo} & \multirow{1}{*}{ZEUS~\cite{zs_audio_ddpm}} & 0.883 & 0.255 & 0.176 & 0.892 & \multirow{1}{*}{0.211} & \multirow{1}{*}{6.41} & 0.21 & 0.30 \\
        TokenFlow~\cite{tokenflow}      &  ZEUS~\cite{zs_audio_ddpm}                  & 0.876 & 0.252 & 0.173 & 0.924 &  0.211&  6.41& 0.20 & 0.27 \\
        RAVE~\cite{rave}                &  ZEUS~\cite{zs_audio_ddpm}                   & 0.885 & 0.251 & 0.170 & 0.881 &  0.211&  6.41& 0.18 & 0.31 \\
        \multicolumn{2}{c|}{RAVE~\cite{rave} $\rightarrow$ Diff-Foley~\cite{diff_foley}} & 0.885 & 0.251 & 0.170 & 0.881 &  0.191&  7.33& 0.16 & 0.35 \\
        \midrule
        \multicolumn{2}{c|}{Delta Denoising Score (DDS)~\cite{dds}} & 0.890 & 0.250 & 0.175 & 0.921 & 0.210 & 5.93 & 0.20 & 0.33\\
        \multicolumn{2}{c|}{\ourmodel (Ours)} & \bf0.903 & \bf0.260 & \bf0.180 & \bf0.956 & \bf{0.226} & \bf{5.55} & \bf{0.23} & \bf{0.42} \\
        \bottomrule
    \end{tabular}}
        \caption{
    \textbf{Comparison to the State-of-the-Art Zero-Shot Video and Audio Editing Models.} 
    We compare our \ourmodel with baselines for zero-shot video~\cite{zhang2023controlvideo,tokenflow,rave,dds} or audio~\cite{zs_audio_ddpm,sdedit,dds} editing on \ourdataa.
    Our evaluation metrics evaluate diverse aspects, including video-only, audio-only, and joint audio-video editing quality.  
    }
    \vspace{\tabmargin}
    \label{tab:sota}
\end{table*}

%% file: figure/human_bar.tex
\begin{figure}[t!]
    \centering
	\includegraphics[width=0.99\linewidth]{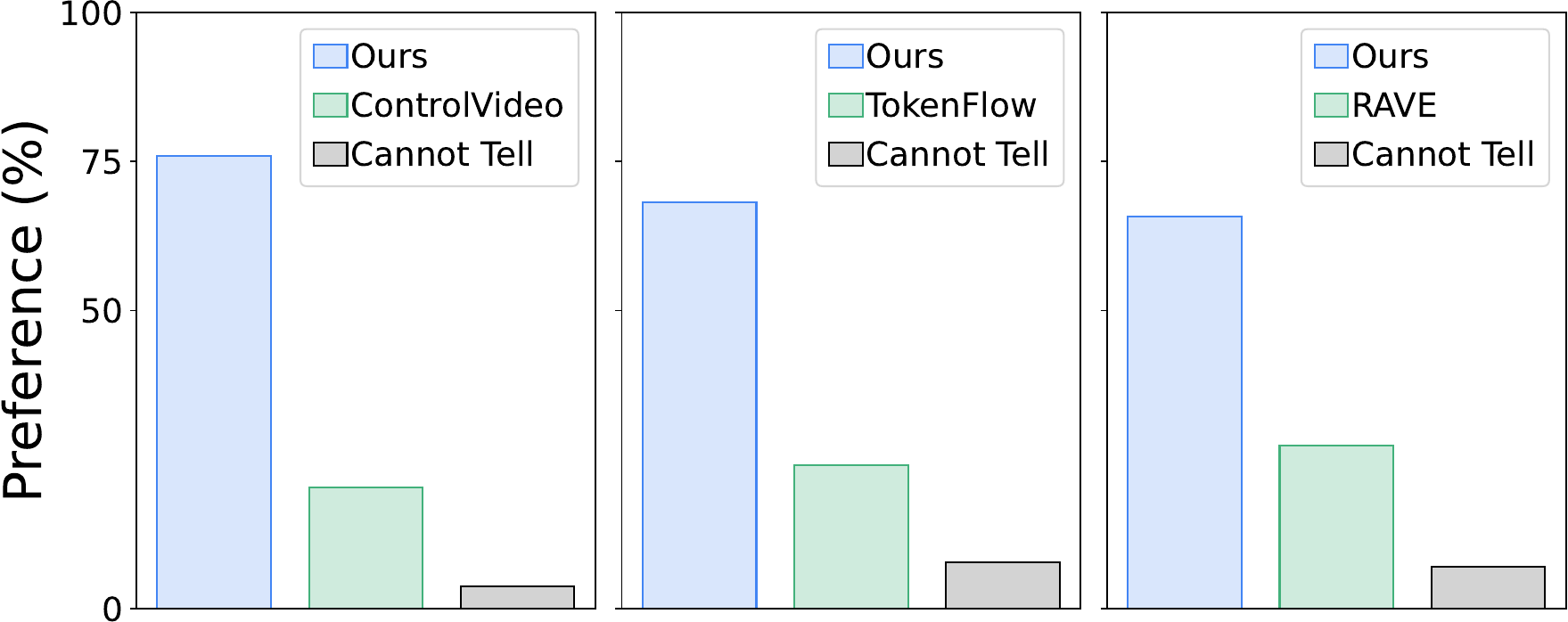}
    \caption{\textbf{Human Evaluation.} Human raters evaluate edited audio and video quality based on alignment with the target text prompt. We report the average human preference rate for each method. All samples are presented in a random order to ensure unbiased assessment.}
    \vspace{\figmargin}
	\label{fig:human}
\end{figure}

%% file: table/sota_oave.tex
\begin{table}[t]
    \centering
    \setlength{\tabcolsep}{2.5pt}
    \resizebox{0.99\linewidth}{!}{
    \begin{tabular}{c|cc|cc|c}
        \toprule
        & \multicolumn{2}{c}{\footnotesize{Video-Only}} & \multicolumn{2}{c}{\footnotesize{Audio-Only}} & \footnotesize{Joint AV} \\
        \cmidrule{2-6}
        \textbf{Method} & \textbf{CLIP-T}$\uparrow$ & \textbf{DINO}$\uparrow$ & \textbf{CLAP}$\uparrow$  & \textbf{LPAPS}$\downarrow$ & \textbf{IB.}$\uparrow$ \\
        \midrule
        ControlVideo~\cite{zhang2023controlvideo}  & 0.261     & 0.892 & 0.245 & 5.91 & 0.28 \\
        TokenFlow~\cite{tokenflow}                 & 0.263     & 0.901 & 0.245 & 5.91 & 0.28 \\
        RAVE~\cite{rave}                           & \bf 0.268 & 0.891 & 0.245 & 5.91 & 0.29 \\
        DDS                                        & 0.260     & 0.930 & 0.240 & 5.58 & 0.28 \\
        \midrule
        \multicolumn{1}{c|}{\ourmodela}            & 0.265     & \bf0.959 & \bf0.250 & \bf5.15 & \bf0.32\\
        \bottomrule
    \end{tabular}}
    \caption{
    \textbf{Comparison on the OAVE Dataset.} 
    We evaluate \ourmodel on the OAVE dataset alongside leading zero-shot audio and video editing models~\cite{zhang2023controlvideo, tokenflow, rave, dds}. 
    The comparison includes video-only, audio-only, and joint audio-video editing metrics. 
    Since OAVE emphasizes audio-image editing, metrics like CLIP-F and AV-Align may be less applicable here.
    }
    \vspace{\tabmargin}
    \label{tab:sota_oave}
\end{table}

%% file: table/abs_single.tex
\begin{table}[t]
    \centering
    \setlength{\tabcolsep}{7pt}
    \resizebox{0.99\linewidth}{!}{
    \begin{tabular}{c c c c}
        \toprule
        \textbf{Configuration} & \textbf{DINO}$\uparrow$ & \textbf{LPAPS}$\downarrow$ & \textbf{AV-Align}$\uparrow$ \\
        \midrule
        Baseline                & 0.921 & 5.93 & 0.33 \\
        + Audio-Only            & 0.921 & 5.60 & 0.36\\
        + Video-Only            & 0.937 & 5.60 & 0.38 \\
        \midrule
        \ourmodel (+AV)                               & \textbf{0.956} & \textbf{5.55} & \textbf{0.42} \\
        \bottomrule
    \end{tabular}}
    \caption{
    \textbf{Cross-Modal vs. Single-Modal Delta Denoising Schemes.} 
    We study \ourmodel with single-modal delta denoising schemes ("Audio-Only" and "Video-Only") and the baseline (\ie \ourmodel without \eqnref{losscmds}) across key metrics about faithfulness, consistency, and audio-visual alignment on \ourdataa. 
    }
    \vspace{\tabmargin}
    \label{tab:abs_single}
\end{table}

%% file: table/abs_select.tex
\begin{table}[t]
    \centering
    \setlength{\tabcolsep}{7pt}
    \resizebox{0.99\linewidth}{!}{
    \begin{tabular}{c c c c}
        \toprule
        \textbf{Configuration} & \textbf{DINO}$\uparrow$ & \textbf{LPAPS}$\downarrow$ & \textbf{AV-Align}$\uparrow$ \\
        \midrule
        Baseline                & 0.921 & 5.93 & 0.33 \\
        Random A/V              & 0.930 & 5.83 & 0.35 \\
        Random A+V              & 0.902 & 6.32 & 0.28 \\
        \midrule
        \ourmodel                               & \textbf{0.956} & \textbf{5.55} & \textbf{0.42} \\
        \bottomrule
    \end{tabular}}
    \caption{
    \textbf{Selecting Positive and Negative Pairs.} 
    We investigate how \ourmodel determines positive and negative pairs without prior knowledge of prompt-relevant patches.
    Similar to the setup in \ourmodel, we randomly sample indices without identifying prompt-relevant patches. In the same modality, patches with the same index across different branches are treated as positive pairs ("Random A/V"). For cross-modal settings (Random A+V), we assume that corresponding audio and video patches are positive pairs.
    }
    \vspace{\tabmargin}
    \label{tab:select}
\end{table}

%% file: figure/fig_sota.tex
\begin{figure*}[t]
    \centering
	\includegraphics[width=0.9\linewidth]{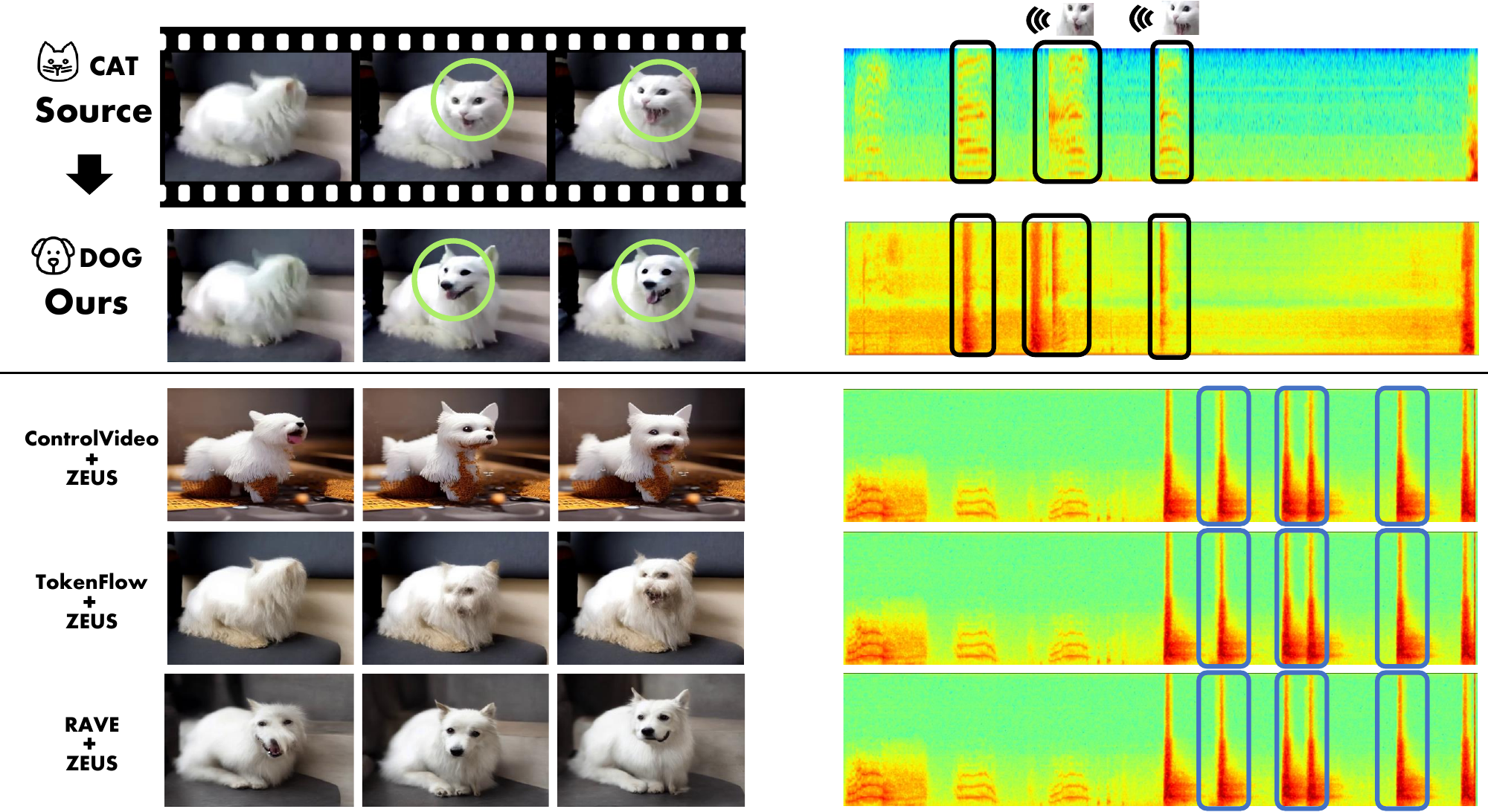}
    \caption{\textbf{Qualitative Zero-Shot Audio-Video Editing Results.}
    We present qualitative results of audio-video editing for a video depicting a transition from "Cat" to "Dog." 
    \ourmodel is compared with video models, including ControlVideo~\cite{zhang2023controlvideo}, TokenFlow~\cite{tokenflow}, and RAVE~\cite{rave}, along with the audio model ZEUS~\cite{zs_audio_ddpm}.
    The \textcolor{green}{\textbf{green}} circles highlight well-aligned motion matches in the video frames, while the black rectangles emphasize precise audio matching.
    The \textcolor{cyan}{\textbf{blue}} rectangles indicate audio artifacts in the competing models, leading to the misalignment between video actions and audio output.
    }
\label{fig:sota}
\vspace{\figmargin}
\end{figure*}

%% file: 5_Conclusion.tex
\vspace{\secmargin}
\section{Conclusions}
\vspace{\secmargin}
\label{sec:conclusions}

In this paper, we introduce \ourmodela, a zero-shot audio-video editing framework developed to address the novel task of synchronized audio-visual editing.
\ourmodel proposes a cross-modal delta denoising scheme that enables synchronized and coherent edits by integrating interactions between audio and video modalities.
To support this task, we curate a benchmark dataset, \ourdataa, which features diverse and challenging audio-visual editing scenarios paired with human-annotated prompts.
We evaluate \ourmodel on \ourdata and the OAVE dataset, demonstrating its superior performance compared to existing single-modality and joint editing baselines.
\ourmodel consistently achieves strong coherence between original and edited content in both modalities, as well as low-level coherence in audio-visual transitions, validating the effectiveness of our cross-modal approach for producing synchronized, high-quality audio-video editing.

\section*{Acknowledgments}
 We thank Md Mohaiminul Islam, Ce Zhang, Yue Yang, Yulu Pan, and Han Yi for their helpful discussions. This work was supported by the Laboratory for Analytic Sciences via NC State University, ONR Award N00014-23-1-2356.

%% file: 6_appendix.tex
\section{Appendix Overview}
Our appendix consists of:
\begin{compactitem}
    \item Details of \ourdataa.
    \item Implementation Details
    \item Human Evaluation Details
    \item Additional Quantitative Results.
\end{compactitem}

\input{supp/fig_heat_map}

\section{Details of \ourdataa}
\paragraph{Category Distribution.}
In \figref{data_src_trg}, we demonstrate the source and target category distributions in the \ourdata dataset to provide a comprehensive overview of its diverse and balanced composition. 
The source categories represent the initial events or objects, while the target categories indicate their corresponding editing events or objects.
\ourdata includes a wide variety of events from animal sounds (\eg \textit{dog}, \textit{cat}, \textit{bird}) to mechanical noises (\eg \textit{machine gun}, \textit{race car}) and environmental effects (\eg \textit{firework}, \textit{water}). 
All categories are well-balanced to ensure that no single category dominates the dataset, which is essential for effective zero-shot audio-video evaluation. 

\paragraph{Mapping of Source and Target Categories.}
In \figref{data_mapping}, We present a heatmap visualizing the count of mappings between source and target categories in the \ourdata dataset. This provides an intuitive understanding of the relationships and transitions from source to target prompts.
Each cell in the heatmap represents the frequency of a specific source-to-target mapping, with darker shades indicating higher counts. 
We note that the mappings include transformations such as \textit{dog} to \textit{lion} and \textit{firework} to \textit{water}, reflecting both logical relationships and imaginative diversity in these pairings. 
These logical and imaginative pairs can fairly and robustly evaluate the effectiveness of audio-video editing tasks.


\begin{figure*}[t!]
    \centering
	\includegraphics[width=0.9\linewidth]{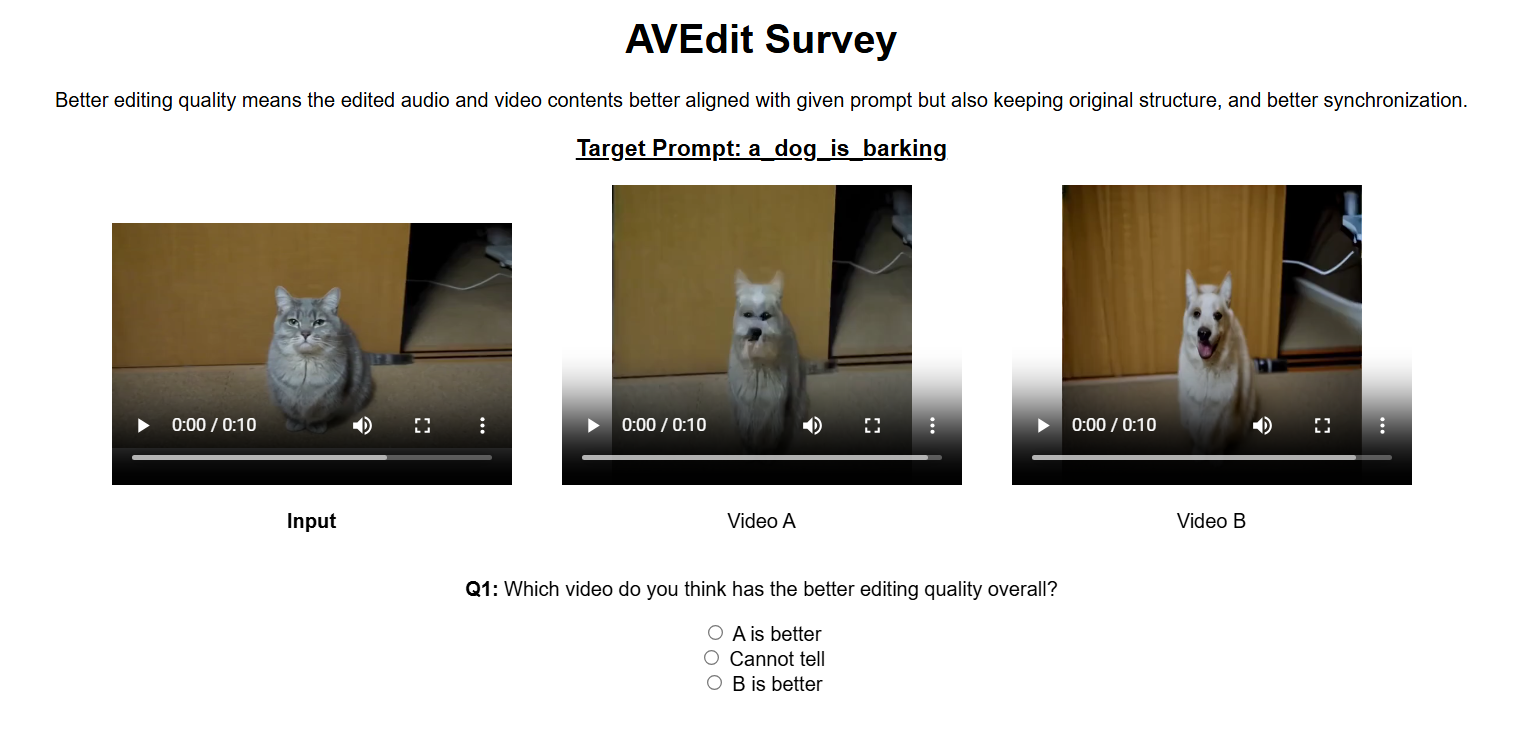}
    \caption{\textbf{Human Evaluation.} Human raters are asked to select the edited video that best aligns with the target prompt. We report the average human preference rate for each method. Note that all samples are presented in a random order.
    }
    \vspace{\figmargin}
	\label{fig:human_web}
\end{figure*}

\section{Implementation Details}
\label{sec:imp}
We use pretrained Stable Diffusion 2.1~\cite{rombach2022highresolution} and AudioLDM2-Large~\cite{audioldm2} as the backbone for video and audio processing, respectively.
Following the setup of RAVE~\cite{rave}, we structure a 10-second video (at $4$ fps) into a $2 \times 2$ grid. At each DDS iteration, the latent frames within each grid are randomly shuffled across different grids.
The optimization process consists of 200 steps in total. 
We optimize the first $15$ steps using only the DDS loss to ensure that the target latent is initially related to the desired editing prompt.
In the remaining steps, we introduce the cross-modal denoising loss $\mathcal{L}_{\text{cmd}}$ by a factor of $10$ for both audio and video.
We adjust the DDS scaling for different phases: for video, the scale is set to 2000 for the first 15 steps and then to 4000 for the remainder.
For audio, it is set to 1000 initially and increases to 5000 after that.
The target latent $\mathbf{z}(\theta)$ is updated using the SGD optimizer with a learning rate of 1, decaying by multiplying 0.99 at each iteration.
We set the threshold both $\tau_{a}$ and $\tau_{v}$ to $0.8$. 
Positive patches are sampled randomly, taking 50\% of the patches where $\tilde{\mathbf{S}}^{a}_{\text{trg}} > \tau$ or $\tilde{\mathbf{S}}^{v}_{\text{trg}} > \tau$ for audio and video, respectively.
For negative sampling, we randomly select 80\% of patches where $\tilde{\mathbf{S}}^{a}_{\text{src}} < \tau$ or $\tilde{\mathbf{S}}^{v}_{\text{src}} < \tau$, from both source and target branches for $\mathbf{H}_{a}^{-}$ and $\mathbf{H}_{v}^{-}$. 
If the number of audio-video patches differs, we randomly drop selected patches to align them.
This entire process takes approximately 20 minutes on a NVIDIA A6000 GPU.

\section{Human Evaluation Details}
As depicted in \figref{human_web}, we conduct a human evaluation to assess the quality of edited audio-video samples based on their alignment with the target prompt.  
Participants are presented with a source (unedited) video and two edited versions generated by different methods (one must come from \ourmodela). 
They are asked to select their preferred sample based on \textit{Which video do you think has the better editing quality overall?} 
For each question, participants can choose one of the two samples or a third option, ``Cannot tell."  
Each subject evaluates five randomly selected video pairs from a pool of 110 comparisons, ensuring a diverse sample set. One sample in each pair is always from \ourmodela, while the other is from a competing method~\cite{zhang2023controlvideo,tokenflow,rave}.  
To prevent bias, all methods remain anonymized during evaluation.  
Our study involves 300 participants recruited via Amazon Mechanical Turk. Results are reported as the average human preference rate for each method, providing insights into the perceived quality of audio-video edits.

\input{supp/abs_threshold}

\section{Additional Quantitative Results}

In \figref{comparison_all}, we present detailed quantitative results evaluating the performance of \ourmodel across different thresholds (\ie $\tau_{v}$ and $\tau_{a}$ in the main draft). 
We report the metrics DINO, LPAPS, and AV-Align, which are highly related to how synchronized edited audio and video are. 
For simplicity, we set $\tau_{v}$ and $\tau_{a}$ \textbf{equally} in these experiments.
These results highlight the impact of different threshold settings on each metric.

\paragraph{DINO and LPAPS.} In \figref{dino} and\figref{lpaps}, these metrics evaluate structural similarity and coherence in visual outputs and perceptual similarity in audio, respectively. The results demonstrate that the score achieves peaks (close to peak) around \textbf{0.8} to suggest the optimal hyper-parameters contributing to aligned audio-video editing.

\paragraph{AV-Align Results.} In \figref{av_align}, the suggested threshold, \textbf{0.8}, also presents the best results in the AV-Align metric to lead the synchronization and coherence between audio and video editing results.

\paragraph{Different Settings for $\tau_v$ and $\tau_a$.} 
The best performance is achieved with $\tau_v = 0.8$ and $\tau_a = 0.7$, yielding the following results: CLIP-F (\textbf{0.905}↑), CLIP-T (\textbf{0.260}↑), Obj. (\textbf{0.180}↑), DINO (\textbf{0.961}↑), CLAP (\textbf{0.229}↑), LPAPS (\textbf{5.41}↓), IB (\textbf{0.24}↑), and AV-Align (\textbf{0.48}↑).  
These results demonstrate the benefits of separately tuning audio and video thresholds to improve overall performance.

\begin{table*}[!t]
\footnotesize
\setlength{\tabcolsep}{4pt} 
\renewcommand{\arraystretch}{0.9} 
\centering
\resizebox{0.8\textwidth}{!}{ 
\begin{tabular}{l|cccc|cc|cc}
    \toprule
    Grid & CLIP-F$\uparrow$ & CLIP-T$\uparrow$ & Obj.$\uparrow$ & DINO$\uparrow$ & CLAP$\uparrow$ & LPAPS$\downarrow$ & IB.$\uparrow$ & Align.$\uparrow$  \\
    \midrule
    2$\times$2 & 0.903 & \textbf{0.260} & \textbf{0.180} & 0.956 & \textbf{0.226} & \textbf{5.55} & \textbf{0.23} & \textbf{0.42}  \\
    3$\times$3 & 0.910 & 0.229 & 0.157 & 0.960 & 0.214 & 5.71 & 0.21 & 0.40  \\
    4$\times$4 & \textbf{0.915} & 0.221 & 0.150 & \textbf{0.961} & 0.211 & 5.65 & 0.21 & 0.40  \\
    \bottomrule
\end{tabular}
}
\caption{\textbf{Grid Design.} Performance comparison across different grid sizes.}
\label{tab:grid_comparison}
\end{table*}

\paragraph{Grid Design Ablation.} 
In \tabref{grid_comparison}, we study different sizes of the grids.
We note that larger grids slightly enhance video temporal consistency (CLIP-F) and visual structure preservation (DINO), while a smaller grid (e.g., $2 \times 2$) yields better visual and audio fidelity (CLIP-T, Obj, CLAP, LPAPS) and synchronization (IB, AV-Align).

\paragraph{Additional Alignment Metric.}
We use the ACC metric~\cite{diff_foley}, which predicts the probability of synchronization, for additional reference.
In \ourdataa, the ACC results$\uparrow$ show that ControlVideo achieves 52.7\%, TokenFlow reaches 45.4\%, and RAVE obtains 55.4\%. In comparison, \ourmodel significantly outperforms these methods with an ACC of \textbf{72.7\%}, highlighting \ourmodela's effectiveness.  
This substantial improvement demonstrates a similar trend of AV-align in the main draft, which ensures better synchronization and alignment of edited content.


\clearpage

%% file: supp/fig_heat_map.tex
\begin{figure*}[t!]
    \centering
	\includegraphics[width=0.95\linewidth]{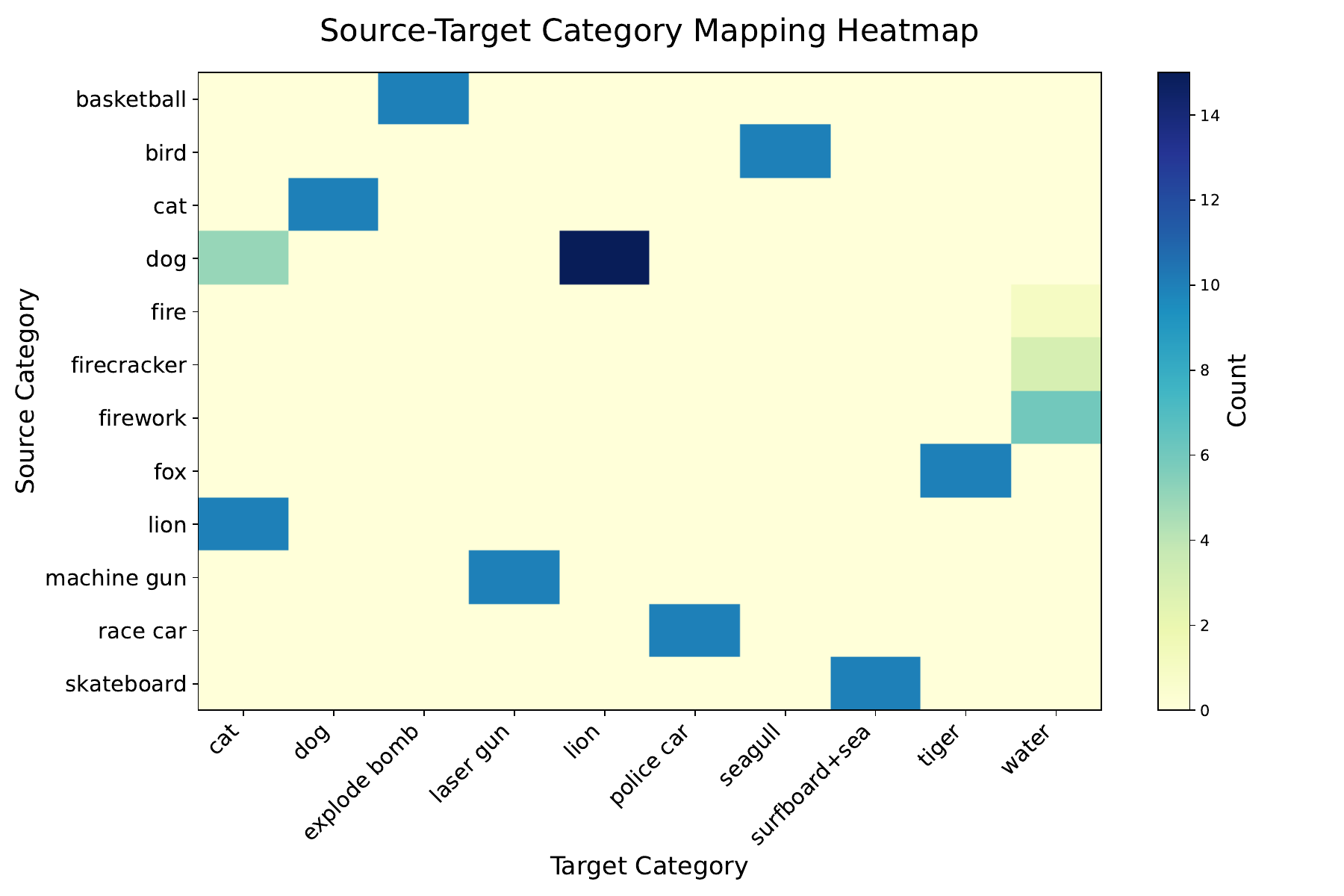}
    \caption{\textbf{Mapping of Source and Target Categories.} 
    This figure summarizes the count of mappings between source and target categories in the dataset. 
    Each cell represents the frequency of a specific source-to-target mapping, providing an intuitive overview of the relationships and transitions present in \ourdataa.
    }
    \vspace{\figmargin}
	\label{fig:data_mapping}
\end{figure*}

%% file: supp/abs_threshold.tex
\begin{figure*}[t]
    \centering
    \begin{subfigure}[b]{0.49\linewidth}
        \centering
        \includegraphics[width=\linewidth]{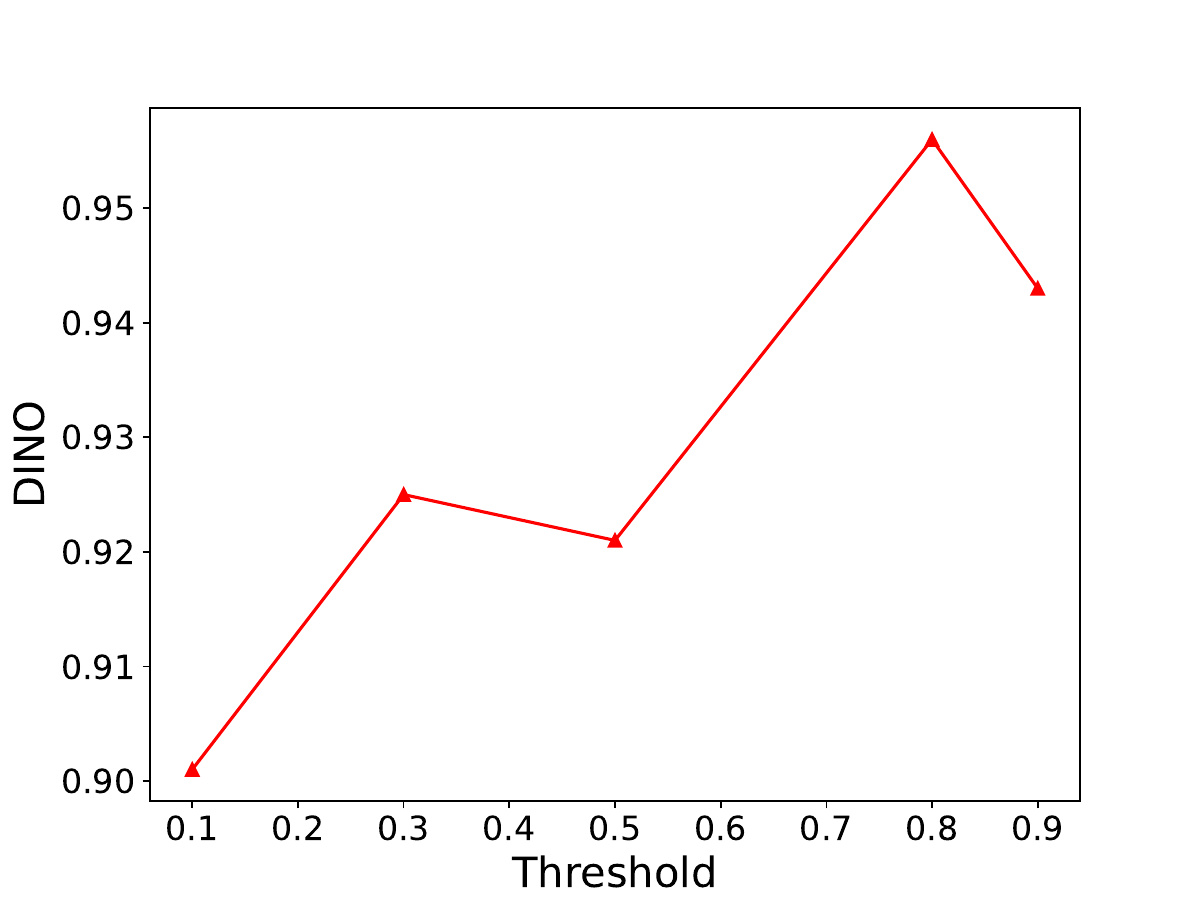}
        \caption{DINO $\uparrow$}
        \label{fig:dino}
    \end{subfigure}
    \hfill
    \begin{subfigure}[b]{0.49\linewidth}
        \centering
        \includegraphics[width=\linewidth]{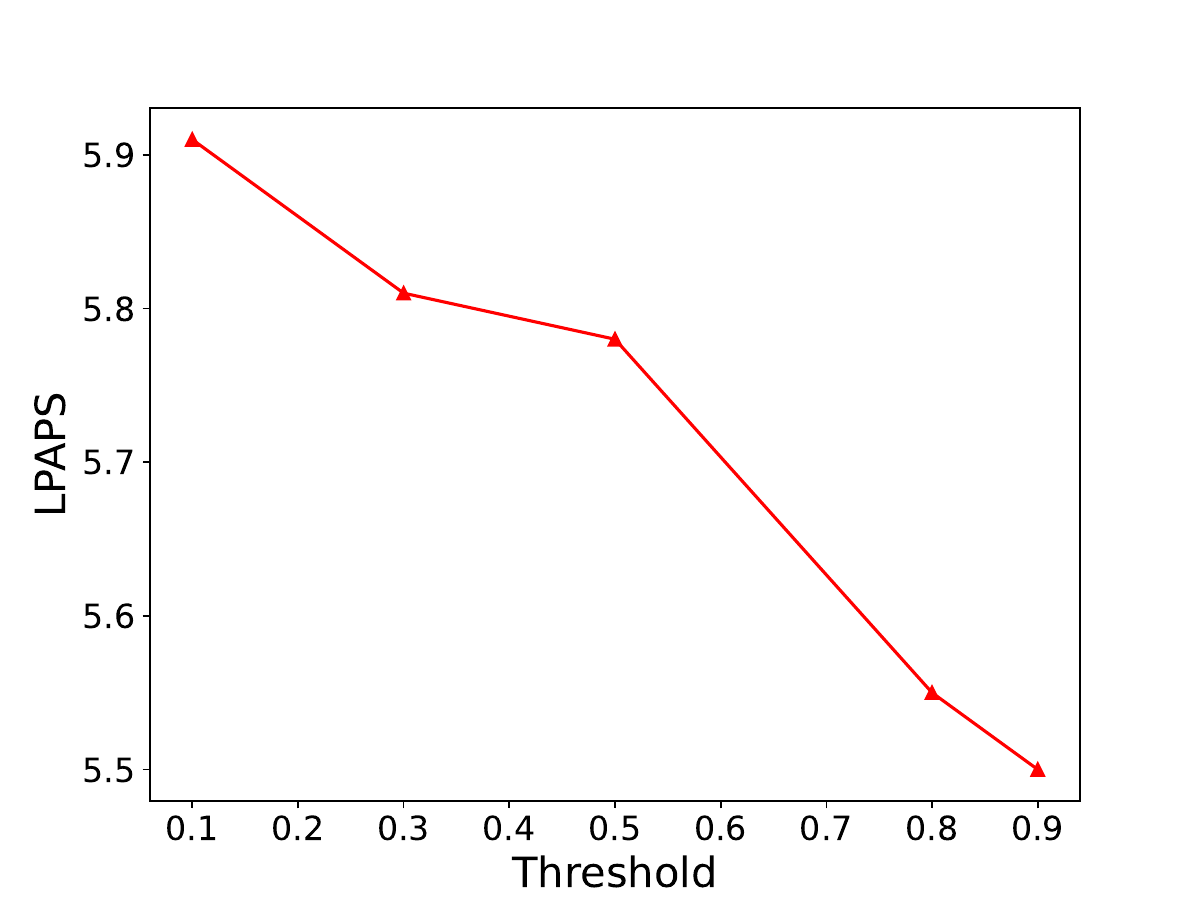}
        \caption{LPAPS $\downarrow$}
        \label{fig:lpaps}
    \end{subfigure}
    \hfill
    \begin{subfigure}[b]{0.49\linewidth}
        \centering
        \includegraphics[width=\linewidth]{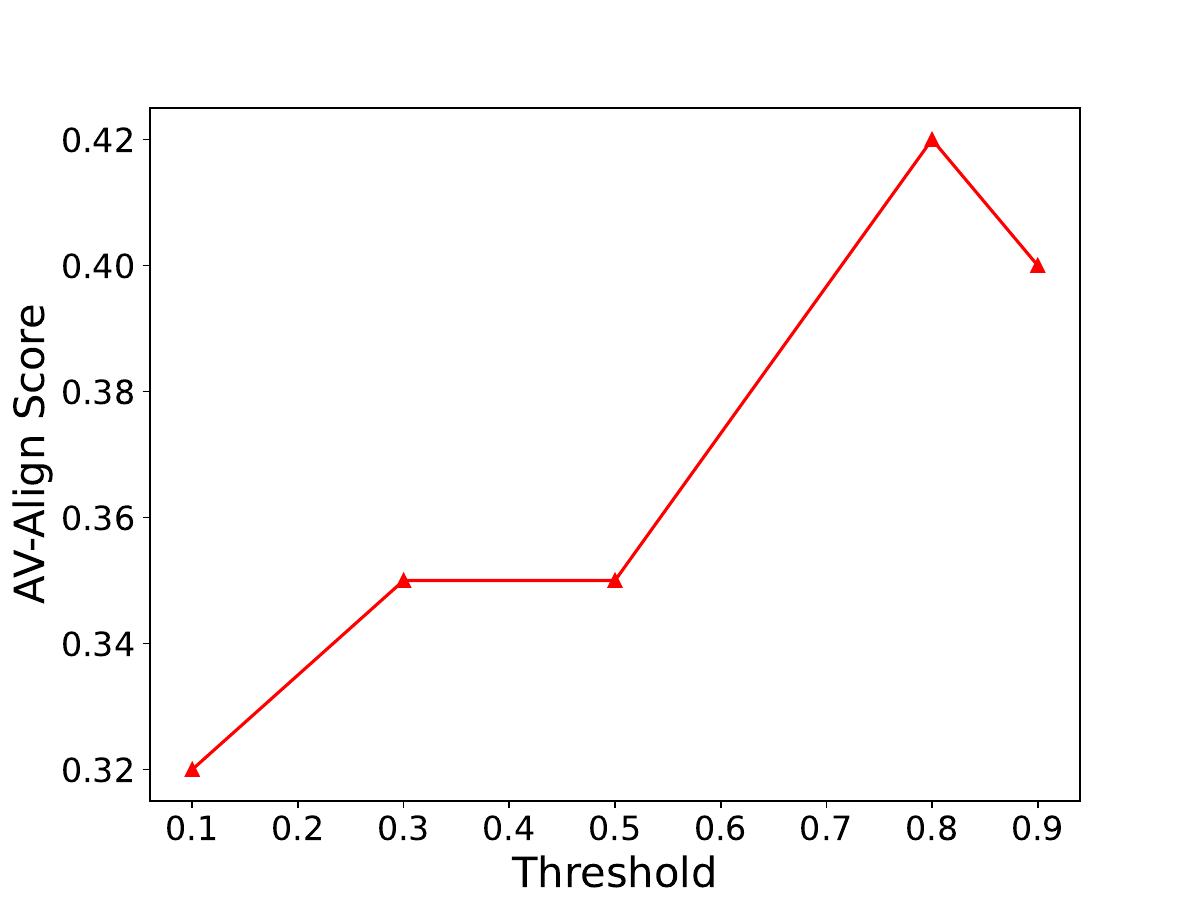}
        \caption{AV-Align $\uparrow$}
        \label{fig:av_align}
    \end{subfigure}

    \caption{\textbf{Impact of the Threshold.} The sub-figures illustrate the performance of DINO, LPAPS, and AV-Align metrics on \ourdata across varying threshold settings, where the threshold decides whether a patch is a prompt-relevant patch (\ie $\tau_{v}$ and $\tau_{a}$ in the main draft). }

    \label{fig:comparison_all}
\end{figure*}